\documentclass[letterpaper, 10 pt, conference]{ieeeconf}  

\IEEEoverridecommandlockouts                              

\title{\LARGE \bf
	RFBNet: Deep Multimodal Networks with Residual Fusion Blocks for RGB-D Semantic Segmentation
}

\author{Liuyuan Deng, Ming Yang, Tianyi Li, Yuesheng He, and Chunxiang Wang
	\thanks{This work is supported by the National Natural Science Foundation of China (U1764264/61873165), Shanghai Automotive Industry Science and Technology Development Foundation (1733/1807), International Chair on automated driving of ground vehicle. Ming Yang is the corresponding author.}
	\thanks{The authors are with Department of Automation, Shanghai Jiao Tong University; Key Laboratory of System Control and Information Processing, Ministry of Education of China, Shanghai, 200240, China (phone: +86-21-34204553; e-mail: MingYang@sjtu.edu.cn).}
}

\usepackage{cite}
\usepackage{flushend}

\usepackage{booktabs}
\usepackage{tabu}
\usepackage{cite}
\usepackage{threeparttable}
\usepackage{amssymb} 

\usepackage{amsmath}

\usepackage{algorithm}
\usepackage{algorithmic}
\usepackage{array}

\usepackage[caption=false,font=footnotesize]{subfig}

\usepackage{xcolor}
\usepackage{xparse}
\usepackage{mathtools}
\NewDocumentCommand{\insquare}{omo}{%
	\begingroup
	\IfValueTF{#1}{%
		\setlength{\fboxsep}{#1}%
	}{%
	}%
	\IfValueTF{#3}{%
		\setlength{\fboxrule}{#3}%
	}{}%
	\ensuremath{\fbox{#2}}
	\endgroup 
}%
\usepackage{pgffor}

\begin{document}

	\maketitle
	\thispagestyle{empty}
	\pagestyle{empty}
	
	\begin{abstract}
		RGB-D semantic segmentation methods conventionally use two independent encoders to extract features from the RGB and depth data. However, there lacks an effective fusion mechanism to bridge the encoders, for the purpose of fully exploiting the complementary information from multiple modalities. This paper proposes a novel bottom-up interactive fusion structure to model the interdependencies between the encoders. The structure introduces an interaction stream to interconnect the encoders. The interaction stream not only progressively aggregates modality-specific features from the encoders but also computes complementary features for them. To instantiate this structure, the paper proposes a residual fusion block (RFB) to formulate the interdependences of the encoders. The RFB consists of two residual units and one fusion unit with gate mechanism. It learns complementary features for the modality-specific encoders and extracts modality-specific features as well as cross-modal features. Based on the RFB, the paper presents the deep multimodal networks for RGB-D semantic segmentation called RFBNet. The experiments on two datasets demonstrate the effectiveness of modeling the interdependencies and that the RFBNet achieved state-of-the-art performance.
	\end{abstract}

	
	\section{INTRODUCTION}
	
	Semantic scene understanding is one of the fundamental tasks for robotics applications, such as precise agriculture\cite{milioto2018real}, autonomous driving\cite{neven2017fast}, semantic mapping and modeling\cite{hermans2014dense,chen20193d}, and localization\cite{stenborg2018long,deng2019semantic}. In recent years, this field has achieved huge progress, thanks to the methodology of convolutional neural network (CNN) based semantic segmentation\cite{shelhamer2017fully,chen2018encoder,zhao2016pyramid}. As depth images provide complementary information to the RGB images, increasing research exploits deep multimodal networks to fuse the two modalities \cite{couprie2013indoor,hazirbas2016fusenet,valada2019self}. This paper investigates the fusion structures of multimodal networks for RGB-D semantic segmentation.
	
	Nowadays, the RGB-D data can be easily obtained by active sensors, e.g., the Microsoft Kinect, or passive sensors, e.g., stereo cameras. The RGB data contain rich appearance information and textural details. Lots of work has been done in semantic segmentation with fully convolutional encoder-decoder networks by using RGB-only data\cite{ romera2017erfnet,chen2018encoder,zhao2016pyramid,deng2019restricted,deng2017cnn,yang2019can}. The depth data provide useful geometric cues which may reduce the uncertainty to segment objects with ambiguous appearance\cite{hazirbas2016fusenet}. It is meaningful and crucial to develop effective models to fuse the two complementary modalities.
	
\begin{figure}[!tb]
		\centering
		\includegraphics[width=0.9\columnwidth]{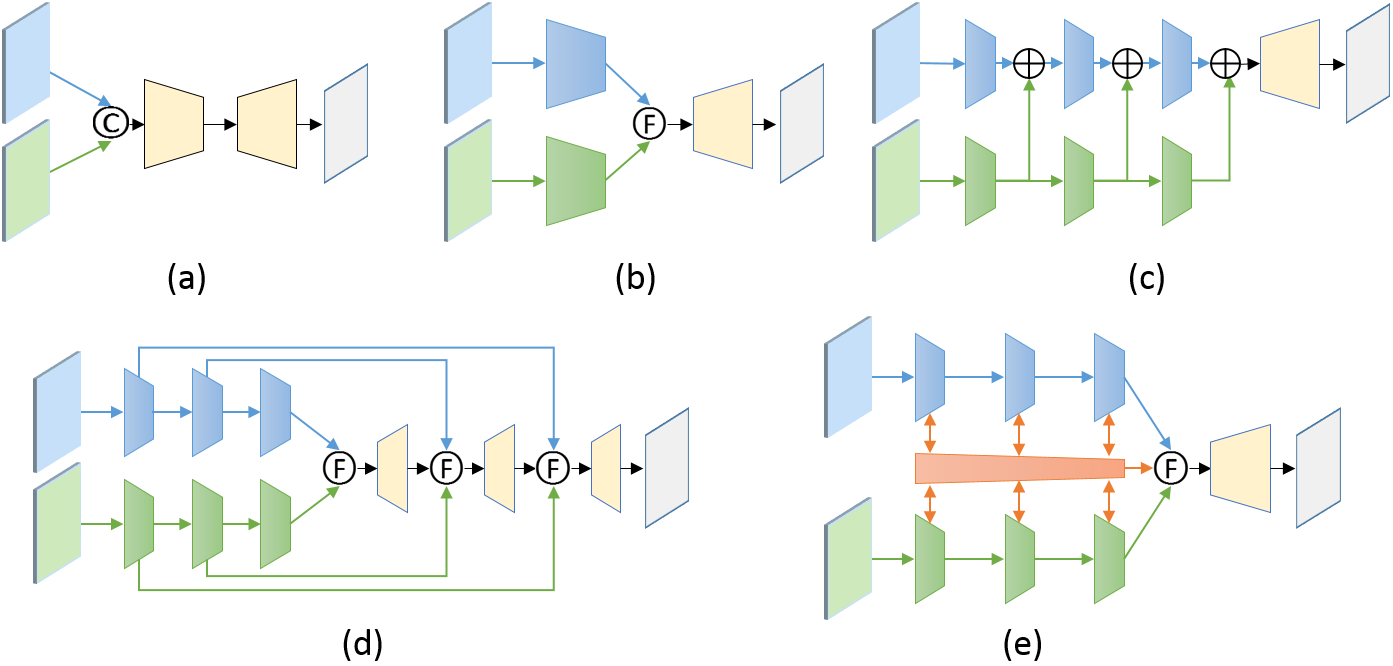}
		\caption{Fusion structures for RGB-D semantic segmentation. The blue color and green color indicate RGB stream and depth stream, respectively. The trapezoids indicate encoder layers and decoder layers. \textcircled{\tiny C} and \textcircled{+} denote the concatenation and summation operations; \textcircled{\tiny F} denotes a combination method. (a) Early fusion (b) Late fusion (c) Fusing multi-level depth features into RGB steam. (d) Top-down multi-level fusion. (e) The proposed bottom-up interactive fusion.}
		\label{fig_fusion_structures}
	\end{figure}

	Many works have shown improvement in semantic segmentation by fusing depth data with RGB data\cite{couprie2013indoor,valada2016deep,cheng2017locality,hazirbas2016fusenet,park2017rdfnet,valada2019self}. Early fusion approaches (see Fig.~\ref{fig_fusion_structures}(a)) simply feed the concatenated RGB and depth channels into a conventional unimodal network\cite{couprie2013indoor}. Such methods may not fully exploit the complementary nature of the modalities\cite{ngiam2011multimodal}. Lots of works turn to the two-stream fusion architecture which processes each modality by a separated and identical encoder and fuses modality-specific features in a single decoder\cite{valada2016deep,valada2019self,cheng2017locality,park2017rdfnet,hazirbas2016fusenet}. The late fusion approaches\cite{valada2016deep, cheng2017locality} (see Fig.~\ref{fig_fusion_structures}(b)) combine the modality-specific features at the end of the two independent encoders with a combination method, e.g., concatenation and element-wise summation. Instead of fusing at early or late stages, hierarchical fusion approaches involve fusing the features at multiple levels. The approaches usually fuse multi-level features from one modality to another modality in the bottom-up path \cite{hazirbas2016fusenet}(see Fig.~\ref{fig_fusion_structures}(c)) or fuse multi-level features in the top-down path\cite{valada2019self, park2017rdfnet} (see Fig.~\ref{fig_fusion_structures}(d)).
	Although these approaches have achieved encouraging results, they do not fully exploit the interdependencies of the modality-specific encoders. It is essential for the encoders to interact and inform each other for reducing the ambiguity in segmentation. How to construct an effective fusion mechanism for bidirectional interaction remains an open problem.

	In this paper, we propose a bottom-up interactive fusion structure to bridge the modality-specific streams with an interaction stream. The structure should address two aspects. First, it should progressively aggregate the information from the modality-specific streams to the interaction stream and extract the cross-model features. Second, it should compute complementary features and feed them to the modality-specific streams without destroying the encoders' ability to extract modality-specific features. The proposed structure is illustrated in Fig.\ref{fig_fusion_structures}(e).
	
	To instantiate this structure, we propose a residual fusion block (RFB) to formulate the interdependencies of the two encoders. The RFB consists of two modality-specific residual units (RUs) and one gated fusion unit (GFU). The GFU adaptively aggregates features from the RUs and generates complementary features for the RUs. The RFB formulates the complementary feature learning as residual learning, and it can extract modality-specific and cross-modal features. With the RFBs, the modality-specific encoders can interact with each other. We build the deep multimodal networks for RGB-D semantic segmentation based on the RFB, which is called RFBNet. And we conduct experiments on two datasets to verify the effectiveness of modeling the interdependencies for RGB-D semantic segmentation.
	
	The main contributions of this paper are summarized as follows:
	\begin{itemize}
		\item We propose the bottom-up interactive fusion structure, which bridges the modality-specific streams, i.e., RGB stream and depth stream, with an interaction stream.
		\item We propose the residual fusion block (RFB) to formulate the interdependencies of the modality-specific streams and build the RFBNet for RGB-D semantic segmentation.
		\item We verify the RFBNet on indoor and outdoor datasets including ScanNet\cite{dai2017scannet} and Cityscapes\cite{cordts2016cityscapes}. The RFBNet constantly outperforms the baselines. Particularly, the model achieves $59.2\%$ mIoU on ScanNet test set.
	\end{itemize}
	
	\section{RELATED WORKS}
	\subsection{Semantic Segmentation}
	Early semantic segmentation methods largely rely on handcrafted features, and use shallow classifiers such as Random Forest and Boosting to predict the class probabilities; then, usually use probabilistic models known as conditional random fields (CRFs) to refine the results\cite{muller2014learning,shotton2009textonboost}. 
	
	In recent years, great progress has been made in this field along with the advance of deep neural networks due to the emerges of large-scale datasets and high-performance graphics processing unit (GPU). FCNs\cite{shelhamer2017fully} successfully improved the accuracy of image semantic
	segmentation by adapting classification networks into fully convolutional networks. The subsequent works\cite{chen2018encoder, li2019gff,zhao2016pyramid} are following this line, including ERFNet\cite{romera2017erfnet} and AdapNet++\cite{valada2019self}. To increase the receptive field and reduce the memory and computational consumption, the encoder-decoder architecture is commonly adopted in these works, in which the encoder gradually reduces the feature maps and captures high-level semantic information, and the decoder recovers the spatial information.
	\subsection{RGB-D data Fusion for Semantic Segmentation}
	The multimodal data fusion has gained long-time attention to exploit the complementary nature of the data of different sources\cite{atrey2010multimodal,ramachandram2017deep}. Early shallow learning methods mainly consider feature-level (early) fusion and decision-level (late) fusion which respectively fusing low-level features and prediction-level features\cite{atrey2010multimodal}. Deep multimodal networks further involve hierarchical fusion or intermediate fusion\cite{ramachandram2017deep,valada2019self} due to the ability of CNNs to learn hierarchical features of the data.
	
	Early fusion can intuitively reuse conventional unimodal semantic segmentation networks\cite{couprie2013indoor,shelhamer2017fully}. For example, Gouprie \textit{et al}.\cite{couprie2013indoor} adapted the multi-scale RGB network of Farabet \textit{et al}.\cite{farabet2012learning} for RGB-D semantic segmentation by concatenating input RGB and depth channels. Late fusion aims to aggregate the high-level features of two modalities using independent networks\cite{gupta2014learning,cheng2017locality,wang2016learning,song2017depth}. Gupta \textit{et al}.\cite{gupta2014learning} concatenated the features extracted by two CNN models from RGB and depth data and classified them with SVM classifier; and they employed a new representation of depth data termed as HHA that encodes horizontal disparity, height above ground and angle with gravity for each pixel. 
	Cheng et al.\cite{cheng2017locality} devised a gated fusion layer to automatically learn the contributions of high-level modality-specific features for an effective combination. Hierarchical fusion enables to combine multimodal features at different layers\cite{hazirbas2016fusenet,jiang2018rednet,park2017rdfnet,valada2019self}.
	FuseNet\cite{hazirbas2016fusenet} fused multi-level depth features into the RGB encoder in the bottom-up path. RedNet\cite{jiang2018rednet} extended FuseNet by additional fusing multi-level features at top-down path. RDFNet\cite{park2017rdfnet} proposed multi-modal feature block and multi-level feature refinement block to fuse multi-level features at top-down path.
	SSMA \cite{valada2019self} proposed a self-supervised model
	adaptation (SSMA) fusion mechanism to combine modality-specific streams and also fused the multi-level features at the top-down path. It achieved state-of-the-art performance on various indoor and outdoor datasets.
	
	The proposed RFBNet also belongs to hierarchical fusion. As a significant difference with existing methods, our approach explicitly formulates the interdependencies of the modality-specific nets, not just aggregating multi-level features.
	
	\subsection{Gate Mechanism}
	Gates are commonly used to regulate the flow of the information\cite{hochreiter1997long,cheng2017locality,li2019gff,srivastava2015training}. Hochreiter \textit{et al}.\cite{hochreiter1997long} used four gates to control the information propagate in and out of the memory cell and Cheng \textit{et al}.\cite{cheng2017locality} used weighted gates to combine features from different modalities automatically. Highway networks\cite{srivastava2015training} used a learned gate mechanism to enable the optimization of very deep networks.
	We also use four gates in the gated fusion unit to regulate the interaction of useful information between modality-specific streams.
	
	\section{Bottom-up Interactive Fusion with Residual Fusion Blocks}
	\begin{figure*}[!tb]
		\centering
		\includegraphics[width=0.7\textwidth]{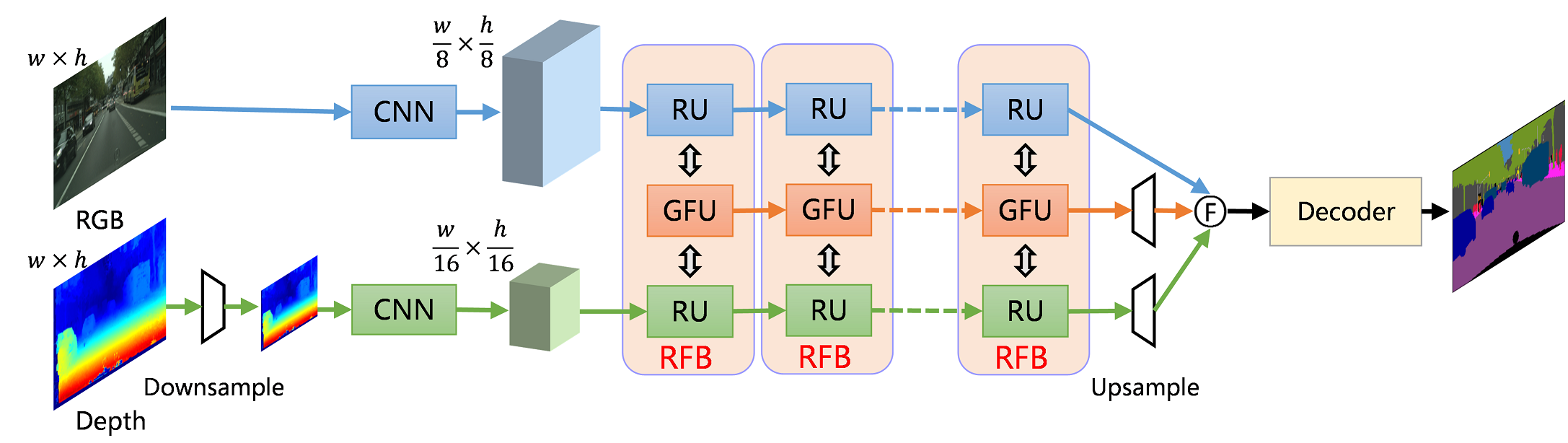}
		\caption{The architecture of the RFBNet. The three bottom-up streams are highlighted by different colors: RGB stream (blue), depth stream (green), and interaction stream (orange). The RFB manages the interaction of the three streams. \textcircled{\tiny F} denotes the combination method.}
		\label{fig_RFBNet}
	\end{figure*}
	
	\subsection{Architecture}
	The architecture of the proposed RFBNet is illustrated in Fig.~\ref{fig_RFBNet}. Besides the RGB steam and depth stream, the architecture introduces an additional interaction stream. The three streams are merged by a combination method such as concatenation, summation, and SSMA block\cite{valada2019self}. Finally, a decoder is appended to compute the predictions. The RFBs are employed at high layers to manage the interaction of the three streams. Specifically, the RFBs are employed at layers after three downsampling operations when the spatial size of the feature map is one eighth that of the input data. Moreover, the spatial size of the interaction stream is the same as that of the depth stream, and the channel dimension is half of that of the depth stream.
	
	\subsection{Shrinking the depth image}
	The architectures with two encoders commonly suffer from large computational and memory consumption. RGB data contains rich appearance and textural details to depict the objects, while depth data contains relatively sparse geometric information to depict the shape of objects. We ease the consumption by shrinking the spatial size of the depth stream. We shrink the depth data by a factor of 2 before inputting into the net, which reduces roughly three-quarters of computation and memory consumption for the depth stream. The depth stream and the interaction stream are upsampled to the same spatial size as the RGB stream before combining. This strategy makes the proposed net slightly faster than the baseline.
	
	
	\subsection{Residual Fusion Block}
	\begin{figure}[!tb]
		\centering
		\includegraphics[width=1.0\columnwidth]{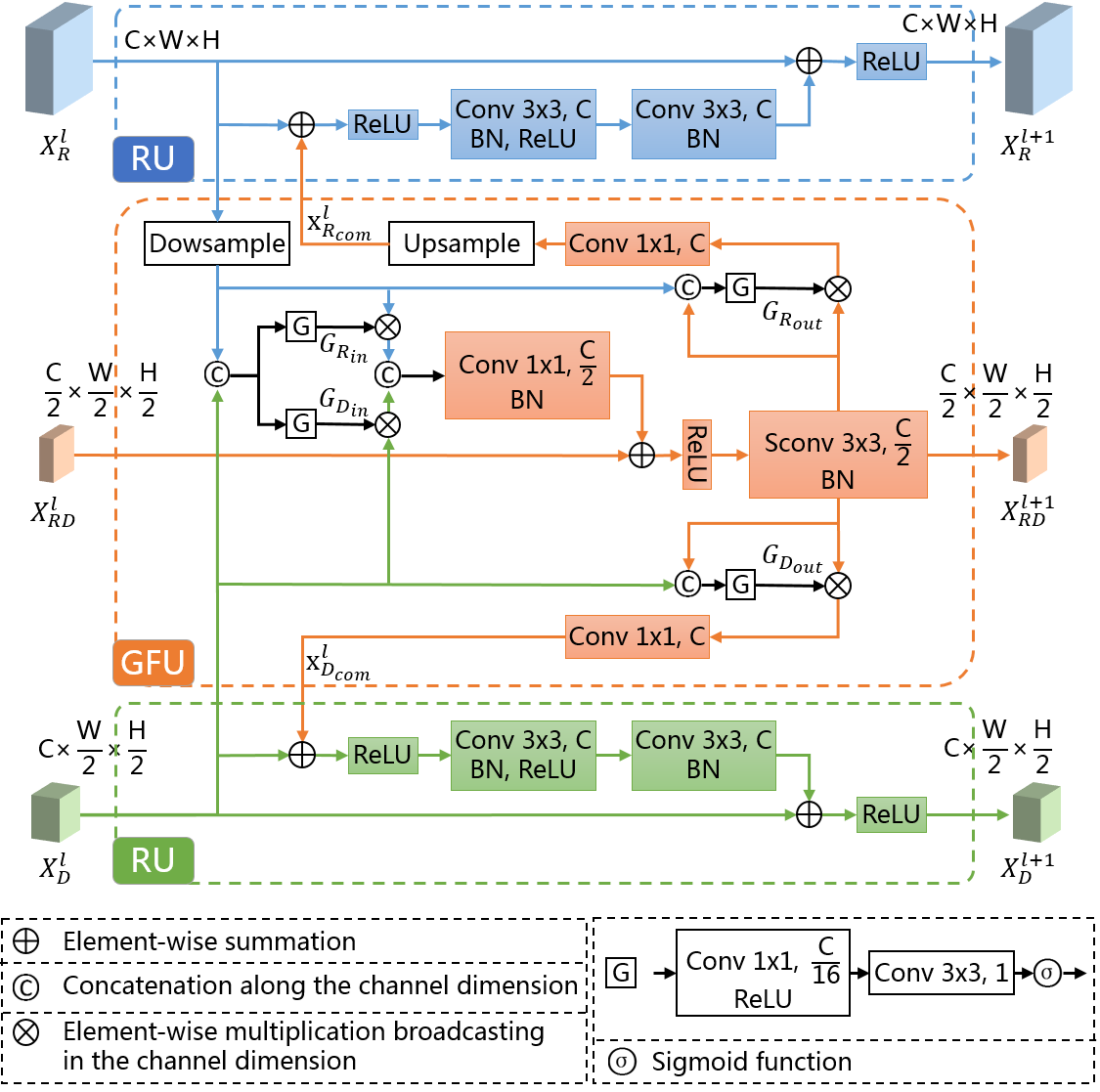}
		\caption{The framework of the residual fusion block (RFB). The block consists of two RUs and one GFU. It manages three streams: RGB, depth, and interaction streams, and formulates the interdependencies of the modality-specific streams. }
		\label{fig_RFB}
	\end{figure}
	The RFB is the basic module to achieve the idea of bottom-up interactive fusion. The RFB consists of two modality-specific residual units (RUs) and one gated fusion unit (GFU). The RU, as the basic unit of ResNet\cite{he2016deep}, is widely used in unimodal networks to learn unimodal features. The RFB learns the modality-specific features based on the RU. We design the GFU to aggregate features from the modality-specific RUs and compute complementary features for the RUs. The framework of the RFB is illustrated in Fig.~\ref{fig_RFB}.

	Given the input RGB, depth, and cross-modal features $\mathrm{x}^l_R\in \mathbb{R}^{C\times H\times W}$, $\mathrm{x}^l_D\in \mathbb{R}^{C\times \frac{H}{2}\times \frac{W}{2}}$, $\mathrm{x}^l_{RD}\in \mathbb{R}^{\frac{C}{2}\times \frac{H}{2}\times \frac{W}{2}}$ and the output features $\mathrm{x}^{l+1}_R\in \mathbb{R}^{C\times H\times W}$, $\mathrm{x}^{l+1}_D\in \mathbb{R}^{C\times \frac{H}{2}\times \frac{W}{2}}$, $\mathrm{x}^{l+1}_{RD}\in \mathbb{R}^{\frac{C}{2}\times \frac{H}{2}\times \frac{W}{2}}$, the RFB is formulated as:
	\begin{equation}
	\mathrm{x}^{l}_{R_{com}}, \mathrm{x}^{l+1}_{RD}, \mathrm{x}^{l}_{D_{com}} = \mathcal{G}(\mathrm{x}^l_R, \mathrm{x}^{l}_{RD}, \mathrm{x}^l_D)
	\label{GFU}
	\end{equation}
	\begin{equation}
	\mathrm{x}^{l+1}_{R}=\mathrm{x}^{l}_{R} + \mathcal{F}_{R}(\mathrm{x}^{l}_{R} + \mathrm{x}^{l}_{R_{com}}, \mathcal{W}^l_R)
	\label{RU1}
	\end{equation}
	\begin{equation}
	\mathrm{x}^{l+1}_{D}=\mathrm{x}^{l}_{D} + \mathcal{F}_{D}(\mathrm{x}^{l}_{D} + \mathrm{x}^{l}_{D_{com}}, \mathcal{W}^l_D)
	\label{RU2}
	\end{equation}
	where $\mathrm{x}^{l}_{R_{com}}$ and $\mathrm{x}^{l}_{D_{com}}$ are the complementary features computed by the GFU denoted as $\mathcal{G}$; $\mathcal{F}_{R}$ and $\mathcal{F}_{D}$ denotes the residual functions of the modality-specific RUs; $\mathcal{W}^l_R$ and $\mathcal{W}^l_D$ are parameters of the RUs. 
	
	From the (\ref{RU1}) and (\ref{RU2}), we can see the modality-specific RUs have the same parameter form with the standard RU\cite{he2016deep}. The difference is that we add complementary features to the input of the RUs. 
	
\begin{figure}[!tb]
		\centering
		\subfloat[]{\includegraphics[height=0.18\columnwidth]{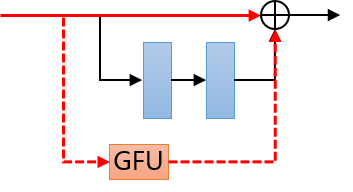}}\hspace{4pt}
		\subfloat[]{\includegraphics[height=0.18\columnwidth]{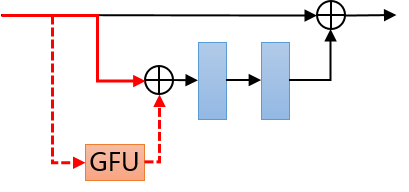}}
		\caption{The flows of the residual and identity mappings when adding complementary features at different points. The GFU acts as a residual function with respect to an identity mapping. The red solid arrow indicates the information flow of identity mapping, while the red dashed arrow indicates the information flow of residual mapping. (a) Adding to the trunk (Point ``T''). (b) Adding to the input of the residual function of the RU (Point ``R'').}
		\label{fig_res}
	\end{figure}

	The RFB formulates the complementary feature learning as residual learning. The GFU acts as a residual function with respect to an identity mapping as illustrated in Fig.~\ref{fig_res}(b). Note that we add the complementary features $\mathrm{x}^{l}_{R_{com}}$ and $\mathrm{x}^{l}_{D_{com}}$ to the inputs of the modality-specific residual functions (denoted as Point ``R'') instead of the trunks of the unimodal streams (denoted as Point ``T''). The different adding points imply different identity mappings as illustrated in Fig.~\ref{fig_res}. The complementary feature directly impacts the modality-specific stream when adding to Point ``T'' (see Fig.~\ref{fig_res}(a)), while it directly impacts the residual function of the modality-specific RU when adding to Point ``R'' (see Fig.~\ref{fig_res}(b)).
	
	
	
	Redundancy, noise, and complementary information exist among different modalities. The GFU explores the underlying complementary relationships in a soft-attention manner via the gate mechanism.
	The GFU contains two input gates and two output gates. The input gates $G_{R_{in}}$, $G_{D_{in}}\in \mathbb{R}^{1\times \frac{H}{2}\times \frac{W}{2}}$ are used to control the unimodal features to flow into the interaction stream, and the output gates $G_{R_{out}}$, $G_{D_{out}}\in \mathbb{R}^{1\times \frac{H}{2}\times \frac{W}{2}}$ are used to regulate the complementary features. The gates are learned by the same network $G(.)$ as shown in the bottom of Fig.~\ref{fig_RFB}. $G(.)$ consists of two convolutional layers with a ReLU layer in between, and a Sigmoid function $\sigma(.)$ to squash values to $[0, 1]$ range. Note that we share the first convolutional layer for input gates to reduce the computational cost. 
	
	The useful information regulated by the input gates is concatenated together, following a $1\times1$ convolutional layer before adding to $\mathrm{x}^l_{RD}$ ($\mathrm{x}^l_{RD}$ is zero for the first RFB). Then we adopt a light-weight depthwise separable convolution (denoted as ``Sconv'' in Fig.~\ref{fig_RFB}) \cite{howard2017mobilenets} to process the cross-modal features in the interaction stream. Finally, the GFU compute the complementary features for the modality-specific RUs regulated by the two output gates $G_{R_{out}}$ and $G_{D_{out}}$.
	
	\subsection{Incooprating with Top-Down Multi-Level Fusion}
\begin{figure}[!tb]
		\centering
		\includegraphics[width=0.5\columnwidth]{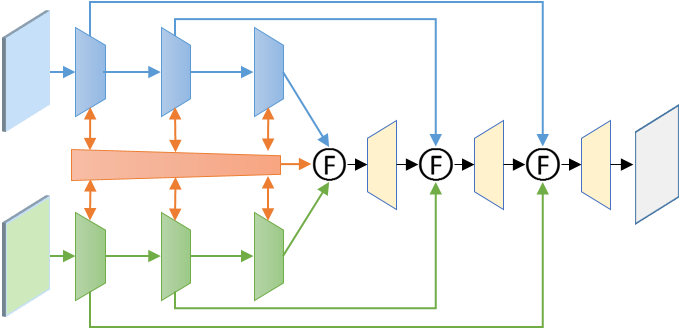}
		\caption{The illustration of a united structure by incorporating bottom-up interactive fusion with top-down multi-level fusion. }
		\label{fig_united}
	\end{figure}

	The proposed bottom-up interactive fusion structure models the interdependencies for the modality-specific encoders. It is orthogonal to the top-down multi-level fusion structure which fuses the encoders features in the top-down path at the decoder stage. The two structures can be incorporated into a united network. We illustrate the structure in Fig.~\ref{fig_united} to give an intuitive understanding. In the experiments, we employ the proposed bottom-up interactive fusion in the SSMA\cite{valada2019self} which adopts the top-down multi-level fusion.
	
	\section{EXPERIMENTAL RESULTS}
	\subsection{Setup}
	\textbf{Datasets}.
	We choose an indoor dataset, i.e, ScanNett\cite{dai2017scannet} and an outdoor dataset, i.e, Cityscapes\cite{cordts2016cityscapes} to evaluate the performance. Each of them provides publicly available training and validation sets as well as an online evaluation server for benchmarking on the test set.
	
	ScanNet is a large-scale indoor scene understanding dataset. It contains $19,466$ samples for training, $5,436$ for validation, and $2,537$ for testing. The RGB images are captured at a resolution of $1296\times 968$ and depth at $640\times 480$. Cityscapes is a large-scale outdoor RGB-D dataset for urban scene understanding. It contains totally $5,000$ finely annotated samples with a resolution of $2048\times 1024$, of which $2,975$ for training, $500$ for validation, and $1,525$ for testing.
	
	\textbf{Backbones.}
	We adopt two unimodal backbones, i.e., AdapNet++\cite{valada2019self} and ERFNet\cite{romera2017erfnet}. AdapNet++ is based on the ResNet-50 model with full pre-activation bottleneck residual units, while ERFNet is a real-time semantic segmentation model based on non-bottleneck factorized residual units. We use the encoder model of the ERFNet (denoted as ERFNetenc) for fast testing and ablation study. A simple bilinear interpolation upsampling layer acts as the decoder in ERFNetenc.
	
	\textbf{Criteria.} We quantify the performance according to the PASCAL VOC intersection-over-union metric (IoU)\cite{everingham2015pascal}.
	
	\subsection{Implementation Details}
	We first built up two-stream fusion networks with the unimodal backbones. We adopt SSMA\cite{valada2019self}, a state-of-the-art method, as the base framework which uses SSMA blocks to combine the two modality-specific streams. The SSMA$_{\textbf{(AdapNet++)}}$ is the same as SSMA model proposed in \cite{valada2019self}. For the SSMA$_{\textbf{(ERFNetEnc)}}$, we use the ERFNetenc to extract two modality-specific features, combine the features with the SSMA block, and use a $1\times 1$ convolutional layer and a bilinear interpolation upsampling layer by a factor of 8 to get the final predictions. 
	To employ our approach, we just replace the corresponding paired RUs with the RFBs for RFBNet$_{\textbf{(AdapNet++)}}$ and RFBNet$_{\textbf{(ERFNetenc)}}$. When employing the RFB for the bottleneck residual units, we feed the output of the first $1\times 1$ layer of the bottleneck RU to the GFU, and add the complementary features to the input of the $3\times 3$ layer of the RU.
	
	The models were implemented using the Tensorflow 1.13.1 and trained on a single 1080Ti GPU. Adam is used for optimization, and ``poly" learning rate scheduling policy is adopted to adjust the learning rate. The weight decay is set to $5e^{-4}$ for the AdapNet++ based models and $1e^{-4}$ for the ERFNetenc based models. The images are resized to a smaller scale for training so that the models can be trained on our 1080Ti GPU. For ScanNet, the images are resized to $640\times 480$; For Cityscapes, the images are resized to $1024\times 512$. We resize the predictions to the full resolution when benchmarking. When training on Cityscapes, we employ a crop of $768\times 384$. 
	
	We first train the unimodal models, then use the trained weights to initialize the encoders of the multimodal models.
	For the AdapNet++ based models, we follow the training procedure of\cite{valada2019self}. We set a mini-batch of 7 for unimodal models, 6 for multimodal models, and 12 for finetuning.
	For the ERFNetenc based models, we use an initial learning rate of $5e^{-4}$. We train $100$K iterations with a mini-batch of 12 for the unimodal models, and 25K iterations with a mini-batch of $9$ for the multimodal models. 
	
	The raw depth data are usually not perfect and have amounts of noise and missing depth values. We perform depth completion\cite{ku2018defense} for the depth images. Moreover, we employ the three-channel HHA encoding\cite{gupta2014learning} for the depth data.
	We employed extensive data augmentations for training, including flipping, scaling, rotation, cropping, color jittering, and Gaussian noise.
	
	\subsection{Results and Analysis}
	\textbf{Benchmarking.} We report the performance benchmarking results on ScanNet and Cityscapes in Table~\ref{tab_scannet_test} and Table~\ref{tab_cityscapes_test}. Note that the test images of the two datasets are not publicly released, and they are used by the evaluation server for benchmarking. From the tables, we can see that the multimodal models have better performance than the unimodal models as expected.
	
	\begin{table}[t]
		\footnotesize 
		\centering
		\caption{Evaluation results on the ScanNet test set.}
		\label{tab_scannet_test}
		\begin{threeparttable}
			\begin{tabular}{c c c}
				\toprule
				Network & Multimodal & mIoU \\
				\midrule
				PSPNet\cite{zhao2016pyramid}    & - & $47.5$		\\
				AdapNet++\cite{valada2019self} & - & $50.3$ 	\\
				3DMV (2d proj)\cite{dai20183dmv} & \checkmark & $49.8$ 	\\
				FuseNett\cite{hazirbas2016fusenet} & \checkmark & $52.1$  		\\
				SSMA\cite{valada2019self} & \checkmark & $57.7$ \\
				RFBNet & \checkmark & $\mathbf{59.2}$	\\
				\bottomrule
			\end{tabular}
		\end{threeparttable}
	\end{table}
	\begin{table}[t]
		\footnotesize 
		\centering
		\caption{Evaluation results on the Cityscapes dataset with different backbones (input image dim: $1024\times 512$).}
		\label{tab_cityscapes_test}
		\begin{threeparttable}
			\begin{tabular}{l c c c}
				\toprule
				Method & Multimodal & mIoU@val & mIoU@test \\
				\midrule
				ERFNetEnc & - & $69.5$ & $67.1$		\\
				SSMA$_{\textbf{(ERFNetEnc)}}$ & \checkmark & $70.8$ & $68.9$	\\
				RFBNet$_{\textbf{(ERFNetEnc)}}$ & \checkmark	& $\mathbf{72.0}$ 	 & $\mathbf{69.7}$ \\
				\midrule
				AdapNet++ & - & $73.4$   & $73.2$		\\
				SSMA$_{\textbf{(AdapNet++)}}$ & \checkmark & $75.0$ & $74.2$ 	\\
				RFBNet$_{\textbf{(AdapNet++)}}$ & \checkmark & $\mathbf{76.2}$	 & $\mathbf{74.8}$ \\
				\bottomrule
			\end{tabular}
		\end{threeparttable}
	\end{table}	
	
	In Table~\ref{tab_scannet_test}, we compare against the top performing models on ScanNet test set. The results are taking from the leaderboard. The proposed RFBNet outperforms other methods, e.g., SSMA\cite{valada2019self}, FuseNet\cite{hazirbas2016fusenet} and 3DMV\cite{dai20183dmv}. Note that the RFBNet and SSMA adopted the same backbone, i.e., AdapNet++, while the SSMA is trained with a batch size of 16 on multiple GPUs with synchronized batch normalization$\footnote{https://github.com/DeepSceneSeg/AdapNet-pp/issues/11}$.
	Still, the RFBNet achieved $1.5\%$ improvement over the SSMA. 
	
	In Table~\ref{tab_cityscapes_test}, we compare RFBNet with base models of different backbones on Cityscapes test set, which shows that the proposed RFBNet constantly outperforms SSMA with different backbones. Note that the accuracy of AdapNet++ and SSMA$_{\textbf{(AdapNet++)}}$ reported in the Table is lower than the official accuracy. This is reasonable because the official models are trained with crops on full resolution images and a larger batch size on multiple GPUs in\cite{valada2019self}.

	We found that the multimodal models improve less on Cityscapes than on ScanNet. We infer the reason is that the depth values of the outdoor data are much noisier and have poorer accuracy than those of the indoor data.
	
\begin{figure*}[!htbp]
		\centering
		\includegraphics[width=0.95\linewidth]{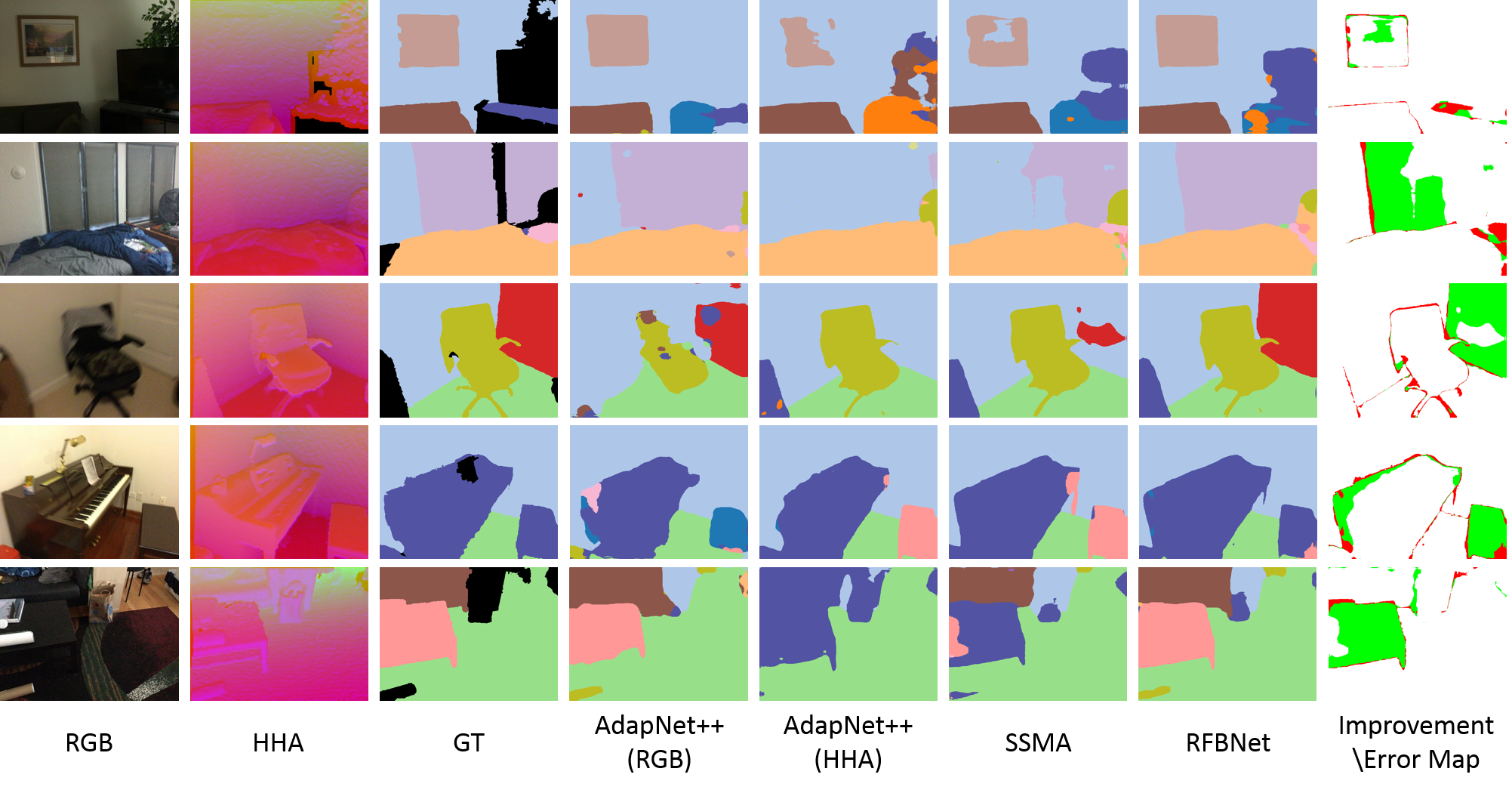}
		\caption{Qualitative results of RFBNet compared with baseline unimodal and multimodal methods on ScanNet dataset. The last column shows the improvement/error map which denotes the misclassified pixels in red and the pixels that are misclassified by SSMA but correctly predicted by RFBNet in green.}
		\label{fig_samples}
	\end{figure*}

\begin{figure*}[!htbp]
		\centering
		\includegraphics[width=1.0\linewidth]{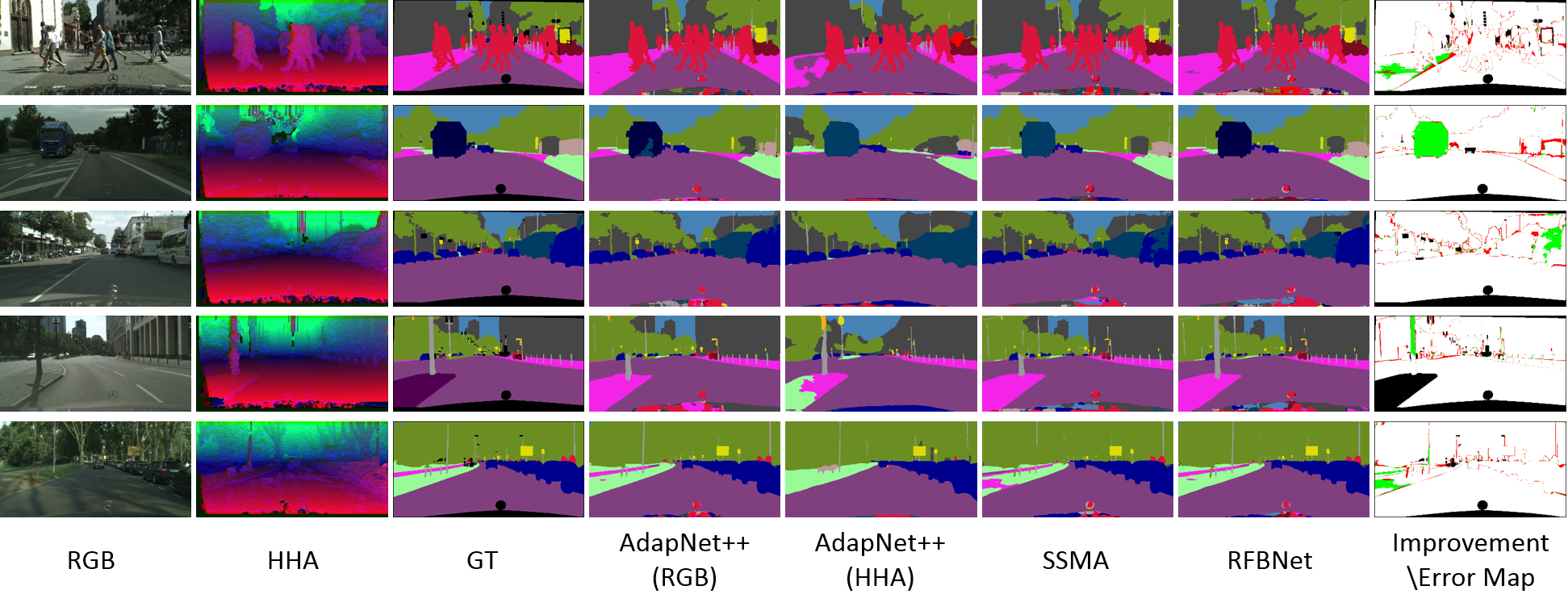}
		\caption{Qualitative results of RFBNet compared with baseline unimodal and multimodal methods on Cityscapes dataset. The last column shows the improvement/error map which denotes the misclassified pixels in red and the pixels that are misclassified by SSMA but correctly predicted by RFBNet in green.}
		\label{fig_samples_cityscapes}
	\end{figure*}

	Some parsing results on ScanNet and Cityscapes are shown in Fig.~\ref{fig_samples} and Fig.~\ref{fig_samples_cityscapes}.
	
	\textbf{Ablation Study.} We perform the ablation study on the ScanNet validation set with the ERFNetenc backbone. In Table~\ref{tab_scannet_erfnetenc_val_depth}, we compare the performance of unimodal models and multimodal models and show how the resolution of the depth data impact the performance. From Table~\ref{tab_scannet_erfnetenc_val_depth}, we can see the multimodal models show a large improvement over unimodal models by more than $4\%$. When shrinking the depth input, the SSMA shows performance decrease by $0.5\%$. We analyze that shrinking the depth can relatively increase the receptive field of depth encoders, which is beneficial for capturing broader context information, but it may lose some geometric details and reduce the spatial representation accuracy. Thus, the performance of SSMA which adopts independent encoders decreases. As the RFBNet bridges the encoders with an interaction stream, both of the unimodal encoders benefit from broader context information. Although losing some geometric details, RFBNet still shows performance improvement by $0.4\%$.
	
	\begin{table}[t]
		\footnotesize 
		\centering
		\caption{Results of ERFNetEnc based models on the ScanNet validation set with different resolutions of depth data.}
		\label{tab_scannet_erfnetenc_val_depth}
		\begin{threeparttable}
			\begin{tabular}{l c c c}
				\toprule
				Method & Input data & Shrink depth & mIoU \\
				\midrule
				ERFNetEnc 		& RGB 		& - & $51.7$	\\
				ERFNetEnc 		& Depth 	& - & $56.7$	\\
				SSMA 	& Multimodal		& $\times$ & $61.6$	\\
				SSMA 	& Multimodal 	& $\checkmark$ & $61.1$ 	\\
				RFBNet 	& Multimodal& $\times$ & $62.2$ 	\\
				RFBNet 	& Multimodal& $\checkmark$ & $\mathbf{62.6}$  \\
				\bottomrule
			\end{tabular}
		\end{threeparttable}
	\end{table}
	\begin{table}[!t]
		\footnotesize
		\caption{Performance of RFBNet$_{\textbf{(ERFNetEnc)}}$ on ScanNet validation set for different settings of the residual fusion block.}
		\label{tab_scannet_erfnetenc_val_rfb}
		\centering
		\begin{threeparttable}
			\begin{tabular}{c c c c}
				\toprule
				G & T & R & mIoU \\
				\midrule
				&			 &			& $61.3$	\\
				$\checkmark$&  			 &			& $61.7$	\\
				$\checkmark$&$\checkmark$&			& $62.0$	\\
				$\checkmark$&			 &$\checkmark$& $\mathbf{62.6}$	\\
				\bottomrule
			\end{tabular}
		\end{threeparttable}
	\end{table}

	We show how the gates and complementary adding points of the RFB impact the performance in Table~\ref{tab_scannet_erfnetenc_val_rfb}. ``G'' means employing gate mechanism to regulate the features. ``T'' and ``R'' means the complementary features are added to the trunk and the input of the residual function of RU, respectively. When ``T'' and ``R'' are disabled, the interaction stream only aggregates features from unimodal encoders but does not compute complementary features for the encoders. From the table, we can see that the performance improves by $0.4\%$ when employing gates. Moreover, enabling ``R'' further improves by $0.9\%$ and outperforms enabling ``T'' by $0.6\%$, which indicates that it is beneficial for the encoders to interact and inform each other.

	\section{CONCLUSION}
	This paper addresses the RGB-D semantic segmentation by explicitly modeling the interdependencies of the RGB stream and depth stream. We proposed a bottom-up interactive fusion structure to bridge the modality-specific encoders with an interaction stream. Specifically, we proposed the residual fusion block to explicitly formulate the interdependences of the two encoders. Experiments demonstrate that the proposed approach achieved considerable improvements by effectively modeling the interdependencies.

	\newpage
	\clearpage
	
	\bibliographystyle{IEEEtran}
	\bibliography{IEEEabrv_lty,bibtex/Multimodal,bibtex/SemanticSegmentation,bibtex/DeepLearning}

\end{document}